# Map Learning with Indistinguishable Locations

Kenneth Basye and Thomas Dean*
Department of Computer Science
Brown University
Box 1910, Providence, RI 02912

## Abstract

Nearly all spatial reasoning problems involve uncertainty of one sort or another. Uncertainty arises due to the inaccuracies of sensors used in measuring distances and angles. We refer to this as *directional* uncertainty. Uncertainty also arises in combining spatial information when one location is mistakenly identified with another. We refer to this as *recognition* uncertainty. Most problems in constructing spatial representations (*maps*) for the purpose of navigation involve both directional and recognition uncertainty. In this paper, we show that a particular class of spatial reasoning problems involving the construction of representations of large-scale space can be solved efficiently even in the presence of directional and recognition uncertainty. We pay particular attention to the problems that arise due to recognition uncertainty.

## 1 Introduction

A *map* is a model of large-scale space used for purposes of navigation. *Map learning* involves exploring the environment, making observations, and then using the observations to construct a map. The construction of useful maps is complicated by the fact that observations involving the position, orientation, and identification of spatially remote objects are invariably error prone. Most studies in map learning have made the simplifying assumption that previously encountered locations can be identified with certainty. In this paper, we consider what happens when you relax that assumption.

In general, local uncertainty accumulates as the product of the distance in generating global esti-mates. One way to avoid this sort of accumulation is to establish strategies such that a robot can discern properties of its environment with certainty. Most existing map learning schemes exploit this sort of certainty in one way or another. The rehearsal strategies of Kuipers [1988] are one example of how a robot might plan to eliminate uncertainty. In situations in which it is not possible to eliminate local uncertainty completely, it is still possible to reduce the effects of accumulated errors to acceptable levels by performing repeated experiments. To support this claim, we describe a map-learning technique based on Valiant's *probably approximately correct* learning model [1984] that, given small $\delta > 0$, constructs a map to answer global queries such that the answer provided in response to any given query is correct with probability $1 - \delta$.

*This work was supported in part by the National Science Foundation under grant IRI-8612644 and by the Advanced Research Projects Agency of the Department of Defense and was monitored by the Air Force Office of Scientific Research under Contract No. F49620-88-C-0132.

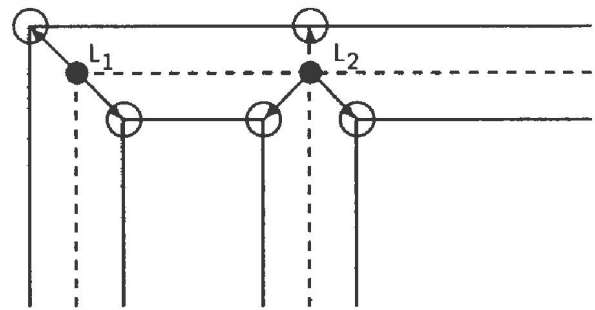

Figure 1: Identifying distinguished locations

## 2 Spatial Modeling

We model the world, for the purposes of studying map learning, as a graph with labels on the edges at each vertex. In practice, a graph will be induced from a set of measurements by identifying a set of distinctive locations in the world, and by noting their connectivity. For example, we might model a city by considering intersections of streets to be distinguished locations, and this will induce a grid-like graph. Kuipers [1988] develops a mapping based on locations distinguished by sensed features like those found in buildings (see Figure 1). Figure 2 shows a portion of a building and the graph that might be induced from it. Levitt [1987] develops a mapping based on locations in the world distinguished by the visibility of landmarks at a distance.

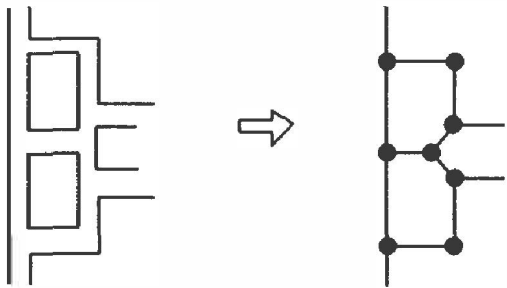

Figure 2: The induced graph of a building

In general, different mappings result in graphs with different characteristics, but there are some properties common to most mappings. For example, if the mapping is built for the purpose of navigating on a surface, the graph induced will almost certainly be planar and cyclic. In what follows, we will always assume that the graphs induced are connected, undirected, and of bounded degree; any other properties will be explicitly noted.

Following [Aleliunas et al., 1979], a graph model consists of a graph, $G = (V, E)$, a set $L$ of labels, and a labeling, $\phi : \{V \times E\} \rightarrow L$, where we may assume that $L$ has a null element $\perp$ which is the label of any pair $(v \in V, e \in E)$ where $e$ is not an edge from $v$. We will frequently use the word *direction* to refer to an edge and its associated label from a given vertex. With this notation, we can describe a path in the graph as a sequence of labels indicating the edges to be taken at each vertex.

If the graph is a regular tessellation, we may assume that the labeling of the edges at each vertex is consistent, i.e., there is a global scheme for labeling the edges and the labels conform to this scheme at every vertex. For example, in a grid tessellation, it is natural to label the edges at each vertex as North, South, East, and West. In general, we do not require a labeling scheme that is globally consistent. You can think of the labels on edges emanating from a given vertex as local directions. Such local directions might correspond to the robot having a compass that is locally consistent but globally inaccurate, or local directions might correspond to locally distinctive features visible from intersections in learning the map of a city.

The robot's activities include moving about in the world and sensing the environment. To model these activities we introduce functions that model the robot's sensors and effectors. A *movement* function is a function from $\{V \times L\} \rightarrow V$. The intuition behind this function is that for any location, one may specify a desired edge to traverse, and the function gives the location reached when the move is executed. A *sensor* function is a function from $V$ to some range of interest. One important sensor function maps vertices to the number of out edges, that is, the degree of the vertex. Another useful function maps vertices to the power set of the set of all labels, $2^L$, giving the possible directions to take from that vertex. We can also partition the set of vertices into some number of equivalence classes and use a function which maps vertices into these classes. We refer to this as a recognition sensor, since it allows the robot to recognize locations. The intuition behind this is that, in some cases, the *local properties* of locations (i.e., the properties that can be discerned of a particular location while situated in that location) enable us to tell them apart. To model uncertainty, we introduce probabilistic forms of these functions. We now develop and explore two kinds of uncertainty that arise in map learning.

### 2.1 Modeling Uncertainty

First, there may be uncertainty in the movement of the robot. In particular, the robot may occasionally move in an unintended direction. We refer to this as *directional* uncertainty, and we model this type of uncertainty by introducing a probabilistic movement function from $\{V \times L\} \rightarrow V$. The intuition behind this function is that for any location, one may specify a desired edge to traverse, and the function gives the location reached when the move is executed. For example, if $G$ is a grid with the labeling given above, and we associate the vertices of $G$ with points $(i, j)$ in the plane, we might define a movement function as follows:

$$\psi((i,j), l) = \begin{cases} (i, j+1) & 70\% \text{ if } l \text{ is North} \\ (i+1, j) & 10\% \text{ if } l \text{ is North} \\ (i-1, j) & 10\% \text{ if } l \text{ is North} \\ (i, j-1) & 10\% \text{ if } l \text{ is North} \\ \ldots & \end{cases}$$

where the "..." indicate the distribution governing movement in the other three directions. The probabilities associated with each direction sum to 1. In this paper, we will assume that movement in the intended direction takes place with probability better than chance.

A second source of uncertainty involves recognizing locations that have been seen before. The robot's sensors have some error, and this can cause error in the recognition of places previously visited; the robot might either fail to recognize some previously visited location, or it might err by mistaking some new location for one seen in the past. We refer to this type of uncertainty as *recognition* uncertainty, and model it by partitioning the set of vertices into equivalence classes. We assume that the robot is unable to distinguish between elements of a given class using only its sensors. In this case the recognition function maps vertices to subsets which are the elements of the partition of the set of vertices. For example, a robot that explores the interior of buildings might use sonar as its primary sensor and use hallway junctions as its distinguished locations. In this case, the robot might be able to distinguish an L junction from a T junction, but might be unable to distinguish between two T junctions. In general, expanding the sensor capabilities of the robot will result in better discrimination of locations, *i.e.*, more equivalence classes, but perfect discrimination will likely be either impractical or impossible. Some locations, however, may be sufficiently distinct that they are distinguishable from all others even with fairly simple sensors. In the model, these locations appear as singleton sets in the partition. We refer to these locations as *landmarks*. We use the term "landmark" advisedly; our landmarks have only some of the usual properties. Specifically, our landmarks are locations that we occupy, not things seen at a distance. They are landmarks because the "view" *from* them is unique. In the following, we make the rather strong assumption that, not only can the robot name the equivalence classes, but it can also determine if a given location is a member of an equivalence class that contains exactly one member (*i.e.*, the robot can identify landmarks).

## 3 Map Learning

For our purposes, a map is a data structure that facilitates queries concerning connectivity, both local and global. Answers to queries involving global connectivity will generally rely on information concerning local connectivity, and hence we regard the fundamental unit of information to be a connection between two nearby locations (*i.e.*, an edge between two vertices in the induced undirected graph). We say that a graph has been *learned completely* if for every location we know all of its neighbors and the directions in which they lie (*i.e.*, we know every triple of the form $\langle u, l, v \rangle$ where $u$ and $v$ are vertices and $l$ is the label at $u$ of an edge in $G$ from $u$ to $v$).

We assume that the information used to construct the map will come from exploring the environment, and we identify two different procedures involved in learning maps: *exploration* and *assimilation*. Exploration involves moving about in the world gathering information, and assimilation involves using that information to construct a useful representation of space. Exploration and assimilation are generally handled in parallel, with assimilation performed incrementally as new information becomes available during exploration.

The problem that we are concerned with in this paper involves both recognition and directional uncertainty with general undirected graphs. In the following, we show that a form of Valiant's probably approximately correct learning is possible when applied to learning maps under certain forms of these conditions.

At any point in time, the robot is facing in a direction defined by the label of a particular edge/vertex pair—the vertex being the location of the robot and the edge being one of the edges emanating from that vertex. We assume that the robot can turn to face in the direction of any of the edges emanating from the robot's location. Directional uncertainty arises when the robot attempts to move in the direction it is pointing. Let $\alpha > 0.5$ be the probability that the robot moves in the direction it is currently pointing. More than 50% of the time, the robot ends up at the other end of the edge defining its current direction, but some percentage of the time it ends up at the other end of some other edge emanating from its starting vertex.

With regard to recognition uncertainty, we assume that the locations in the world are of two kinds, those that can be distinguished, and all others. That is, there is some set of landmarks, in the sense explained above, and all other locations are indistinguishable. We model this situation using a partitioning $W$ of $V$ and assuming that we have a sensor function which maps $V$ to $W$. $W$ consists of some number of singletons and the set of all indistinguishable elements. We further assume that a second sensor function allows us to determine whether the current location is or is not a landmark. For convenience, we define $D$ to be the subset of $V$ consisting of all and only landmark vertices, and $I$ to be the subset of $V$ consisting of all and only non-landmark vertices. We refer to this kind of graph as a *landmark graph*. We define the *landmark distribution parameter*, $r$, to be the maximum distance from any vertex in $I$ to its nearest landmark (if $r = 0$, then $I$ is empty and all vertices are landmarks). We say that a procedure learns the *local connectivity within radius $r$* of some $v \in D$ if it can provide the shortest path between $v$ and any other vertex in $D$ within a radius $r$ of $v$. We say that a procedure learns the *global connectivity of a graph $G$ within a constant factor* if, for any two vertices $u$ and $v$ in $D$, it can provide a path between $u$ and $v$ whose length is within a constant factor of the length of the shortest path between $u$ and $v$ in $G$. The path will be constructed from paths found between locally connected landmarks (see Figure 3).

In the following, we assume that the probability of the robot guessing that it did traverse a path $p$ given that it actually did traverse $p$ is $\gamma$, that $\gamma > \frac{1}{2} + \epsilon$ where $\epsilon$ is positive, and that the robot knows these two facts. The answers to these guesses might be arrived at by various means. First, some monitoring of the robot's movement mechanisms could provide an indication of the quality of the traversal. Any *a priori* information about the path could be used to

9

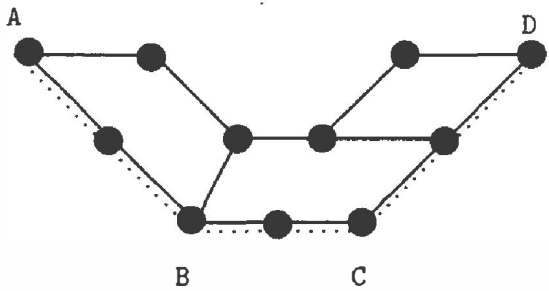

Figure 3: The path found between landmarks $A$ and $D$

provide the answer, and some information regarding features seen in the previous exploration steps might be useful here as well.

We begin by showing that the multiplicative error incurred in trying to answer global path queries can be kept low if the local error can be kept low, that the transition from a local uncertainty measure to a global uncertainty measure does not increase the complexity by more than a polynomial factor, and that it is possible to build a procedure that directs exploration and map building so as to answer global path queries that are accurate and within a small constant factor of optimal with high probability.

**Lemma 1** *Let $G$ be a landmark graph with distribution parameter $r$, and let $c$ be some integer $> 2$. Given a procedure that, for any $\delta_l > 0$, learns the local connectivity within $cr$ of any landmark in $G$ in time polynomial in $\frac{1}{\delta_l}$ with probability $1 - \delta_l$, there is a procedure that learns the global connectivity of $G$ with probability $1 - \delta_g$ for any $\delta_g > 0$ in time polynomial in $\frac{1}{\delta_g}$ and the size of the graph. Any global path returned as a result will be at most $\frac{c}{c-2}$ times the length of the optimal path.*

**Proof:** Let $m$ be the length of the longest answer we might have to provide to a global query. Then the probability of correctness for any global answer obeys

$$p(\text{correct answer}) \geq (1 - \delta_l)^m$$

A simple expansion gives

$$(1 - \delta_l)^m = 1 - m\delta_l + E \geq 1 - m\delta_l$$

because $E \geq 0$. Thus, ensuring that every $\delta_l = \delta_g/m$ will ensure that

$$p(\text{correct answer}) \geq 1 - \delta_g$$

We use the local procedure on every distinguishable vertex in the graph and the resulting representation is sufficient to provide a path between any two distinguishable vertices. Note that we do not have to know $|V|$ in order to calculate $\delta_l$, only the length of the longest answer expected. The proof that the resulting paths are within a constant factor of optimal appears in [Basye *et al.*, 1989].

**Lemma 2** *There exists a procedure that, for any $\delta_l > 0$, learns the local connectivity within $cr$ of a vertex in any landmark graph with probability $1 - \delta_l$ in time polynomial in $\frac{1}{\delta_l}$, $\frac{1}{2\gamma - 1}$ and the size of $G$, and exponential in $r$.*

**Proof:** The learning algorithm can be broken down into three steps: a landmark identification step in which the robot finds and identifies a set of landmarks, a candidate selection step in which the robot finds a set of candidates for paths in $G$ connecting landmarks, and a candidate filtering step in which the robot determines which of those candidates actually correspond to paths in $G$. In order to prove the lemma, landmark identification has to succeed in identifying all landmarks in $G$ with high probability, candidate selection has to find all paths (or at least all of the shortest paths) between landmarks with high probability, and candidate filtering has to determine which of the candidates correspond to paths in $G$ with high probablity. Let $1 - \delta_i$, $1 - \delta_s$, and $1 - \delta_f$ correspond, respectively, to the probabilities that the three steps succeed in performing their associated tasks. We will consider each of the three steps in turn.

The first step is easy. The robot identifies all the landmarks in $G$ with probability $1 - \delta_i$ by making a random walk whose length is polynomial in $\frac{1}{\delta_i}$ and the size of $G$. A more sophisticated exploration might be possible, but a random walk suffices for polynomial-time performance.

Having identified a set of landmarks, the robot has to try all paths of length $r$ or less starting from each identified landmark. If $d$ is the maximum degree of any vertex in $G$, then there can be as many as $d^r$ paths of length $r$ or less starting from any vertex in $G$. This requires than an exhaustive search will be exponential in $r$. Since we expect that $r$ will generally be small, this "local" exponential factor should not be critical. For each landmark, the robot tries some number of paths of length $r$ trying to connect other landmarks within a radius $r$. Again, a simple coin-flipping algorithm will do for our purposes. Starting from a landmark $A$, the robot chooses randomly some direction to follow, it records that direction, and then attempts to follow that direction. It continues in this manner until it has taken $r$ steps. If it encounters one or more landmarks (other than $A$), then it records the set of directions attempted as a candidate path. The resulting candidates look like:

$$A_{out_0}, {}_{in_1}X_{out_1}, \ldots, {}_{in_{k-1}}X_{out_{k-1}}, {}_{in_k}B$$

where $B$ is the landmark observed on a path starting from $A$, and the notation ${}_{in}X_{out}$ indicates that the robot *observed* itself entering a vertex of type $X$ on the arc labeled *in* and it *observed* itself attempting to leave on the arc labeled *out*. The probability that the robot will traverse a particular path of length $r$ on any given attempt is $\frac{1}{d^r}$. The probability that the robot will traverse the path of length $r$ that it attempts to traverse is $\alpha^r$. Since the robot records only those paths it attempts, it has to make enough attempts so that with high probability it records all the paths. The probability that the robot will record any given $r$-length path on $n$ attempts starting at $A$ is:

$$1 - \left[1 - \left(\frac{\alpha}{d}\right)^r\right]^n$$

In order to ensure that we record all such paths with probability $1 - \delta_s$ we have to ensure that:

$$1 - \delta_s \leq \left[1 - \left[1 - \left(\frac{\alpha}{d}\right)^r\right]^n\right]^{d^r}$$

Solving for $n$ we see that the robot will have to make a number of attempts polynomial in $\frac{1}{\delta_s}$ and exponential in $r$.

Candidate filtering now proceeds as follows for each candidate path. The robot attempts to traverse the path, and, if it succeeds, it guesses whether or not it did so correctly. A traversal of the path that was correct indicates that the path really is in $G$. With directional uncertainty, it is possible that although the traversal started and ended at the right locations and seemed to take the right direction at each step, the path actually traversed is not the one that was attempted. This results in a "false positive" observation for the path in question. The purpose of the guess after a traversal is to distinguish false positives from correct traversals. For each traversal that succeeds, we record the answer to the guess, and we keep track of the number of positive and negative answers. After $n$ traversals and guesses, if the path really is in $G$, we expect the number of positive answers to be near $n\gamma$. We use $n/2$ as the threshold, and include only paths with more than $n/2$ positive answers in our representation. By making $n$ sufficiently large, we can assure that this filtering accepts all and only real paths with the desired probability, $1 - \delta_f$. We now consider the relation between $n$ and $\delta_f$.

The entire filtering step will succeed with global probability $1 - \delta_f$ if we ensure that each path is correctly filtered with some local probability, which we will call $\delta_{fl}$. An argument similar to the one used in the proof of Lemma 1 shows that the local probability is polynomial in the global probability, $\delta_f$, $d$, and the size of $G$, and exponential in $r$. We now show that the number of traversals, $n$, is polynomial in $\frac{1}{\delta_{fi}}$ and $\frac{1}{(2\gamma-1)}$.

As mentioned above, in $n$ traversals we expect about $n\gamma$ positive answers if the path is really in $G$. We use $n/2$ as a threshold, and we wish to ensure that this includes all and only real paths. We therefore consider the probability that we will get $n/2$ or fewer positive responses even though the path is really in $G$. This case covers the possibility of wrongly including a path; the analysis covering wrongly excluding a path is similar. We assume that the number of positive responses will be normally distributed about a mean of $n\gamma$ if the path is real. The probability of making an error is the probability that the number of positive answers will deviate from this mean enough to fall below $n/2$. If $X$ is the number of positive responses we get, then

$$P(error) \leq P(|X - n\gamma| \leq n(\gamma - \frac{1}{2})) \leq \frac{\gamma(1-\gamma)}{n(\gamma - \frac{1}{2})^2}$$

Replacing $\gamma(1 - \gamma)$ with $\frac{1}{4}$, we have

$$P(error) \leq \frac{1}{n(2\gamma - 1)}$$

which will be less than $\delta_{fi}$ provided that

$$n \geq \frac{1}{\delta_{fi}} \frac{1}{(2\gamma - 1)^2}$$

**Theorem 1** *It is possible to learn the global connectivity of any landmark graph with probability $1 - \delta$ in time polynomial in $\frac{1}{\delta}$, $\frac{1}{2\gamma - 1}$, and the size of $G$, and exponential in $r$.*

Theorem 1 is a simple consequence of Lemma 1 and 2. It has an immediate application to the problem of learning the global connectivity of a graph where all the vertices are landmarks. In this case, the parameter $r = 0$, and we need only explore paths of length 1 in order to establish the global connectivity of the graph.

**Corollary 1** *It is possible to learn the connectivity of a graph $G$ with only distinguishable locations with probability $1 - \delta$ in time polynomial in $\frac{1}{\delta}$, $\frac{1}{1-2\alpha}$, and the size of $G$.*

## 4  Discussion

The proof in the previous section relies on the assumption that the robot knows it can identify the correct execution of a set of instructions specifying a path with probability better than $\gamma$. Knowing the value of $\gamma$ enables the robot to determine how many experiments it must perform in order to construct a map that is correct with probability $1 - \delta$. The intuition behind this is that, in generating each candidate path in the initial exploration phase, the robot also compiles a set of observations (*e.g.*, local features, distances traveled, and angles turned) to be used as expectations during the candidate filtering step. The expectations are used to rule out situations in which the robot fails to correctly execute the instructions in the candidate path.

We have also considered the case in which movement in the intended direction takes place with probability better than chance, and that, upon entering

11

a vertex, the robot knows with certainty the local name of the edge upon which it entered. We call the latter ability *reverse movement certainty*. In traversing an edge the robot will not know that it has ended up at some unintended location, but it will know what direction to follow in trying to return to its previous location.

With the assumption of reverse movement certainty, we have additional information that we can bring to bear on distinguishing successes from failures. As mentioned earlier, in the initial exploration phase, the robot generates a set of candidate paths from observations, where each candidate is of the form:

$$p = A_{out_0}, {}_{in_1}X_{out_1}, \ldots, {}_{in_{k-1}}X_{out_{k-1}}, {}_{in_k}B.$$

Given reverse movement certainty, the set of directions indicated by the labels $in_k, in_{k-1}, \ldots, in_1$ are guaranteed to describe a path from $B$ to $A$ in $G$. What we have to determine is whether or not the set of directions $out_0, out_1, \ldots, out_{k-1}$ describe a path from $A$ to $B$ in $G$.

To make this determination, the robot runs a set of experiments. In each experiment, the robot tries to follow the directions indicated by $in_k, in_{k-1}, \ldots, in_1$, and it keeps track of the number of *hits*: experiments in which it observes the sequence of labels $out_{k-1}, out_{k-2}, \ldots, out_0$ on entering vertices. If $p$ is a path, then in $n$ experiments the expected number of hits is $\alpha^d n$. If $p$ is not a path, then the expected number of hits is $\alpha^{d-1}(1-\alpha)n$ or less depending upon how many movement errors were made in the original traversal. It is this separation between $\alpha^d$ and $\alpha^{d-1}(1-\alpha)$ that we exploit in determining whether or not a candidate path is actually a path in $G$.

Given the notion of global connectivity defined above, no attempt is made to *completely learn* the graph (i.e., to recover the structure of the entire graph). It is assumed that the indistinguishable vertices are of interest only in so far as they provide directions necessary to traverse a direct path between two landmarks. But it is easy to imagine situations where the indistinguishable vertices and the paths between them are of interest. For instance, the indistinguishable vertices might be partitioned further into equivalence classes so that one could uniquely designate a vertex by specifying its equivalence class and some radius from a particular global landmark (e.g., the bookstore just across the street from the Chrysler building). In [Basye et al., 1989], we show how our approach can be applied to completely learn the graph by first completely learning local neighborhoods of each landmark.

## 5 Related Work

Kuipers defines the notion of "place" in terms of a set of related visual events [Kuipers, 1978]. This notion provides a basis for inducing graphs from measurements. In Kuipers' framework [1988], locations are arranged in an unrestricted planar graph. There is recognition uncertainty, but there is no directional uncertainty (if a robot tries to traverse a particular hall, then it will actually traverse that hall; it may not be able to measure exactly how long the hall is, but it will not mistakenly move down the wrong hall). Kuipers goes to some length to deal with recognition uncertainty. To ensure correctness, he has to assume that there is some reference location that is distinguishable from all other locations. Since there is no directional uncertainty, any two locations can be distinguished by traversing paths to the reference location. Given a procedure that is guaranteed to uniquely identify a location if it succeeds, and succeeds with high probability, we can show that a Kuipers-style map can be reliably probably almost always usefully learned using an analysis similar to that of Section 3. In fact, we do not require that the edges emanating from each vertex be labeled, just that they are cyclically ordered.

Levitt *et al* [1987] describe an approach to spatial reasoning that avoids multiplicative error by introducing local coordinate systems based on landmarks. Landmarks correspond to environmental features that can be acquired and, more importantly, reacquired in exploring the environment. Given that landmarks can be uniquely identified, one can induce a graph whose vertices correspond to regions of space defined by the landmarks visible in that region. If the identification and reaquisition of landmarks is guaranteed, then the problem involves neither recognition nor movement uncertainty. (The robot will not cross the line between one pair of landmarks thinking that it has crossed the line between some other pair of landmarks.) Of course, in an outdoor environment, landmark identification and reaccquistion are both situation and aspect dependent (i.e., the current environmental factors and the direction from which the robot views a scene influence both recognition and reacquisition). In such cases, some amount of recognition uncertainty will certainly manifest itself.

The sensor and movement functions presented in this paper are primarily useful for the analysis of map learning in which sensing is local and the environment restricts the choice of movement at any given location to a small number of options (e.g., office-like environments). In Kuipers' work landmarks are few and may be difficult to identify, but movement errors are nonexistent. In Levitt's work, landmarks are plentiful and easy to identify, metric information is error prone, but the uncertainty is small. We are currently considering how to apply our methods to the type of problem discussed in Levitt's work, but there are a number of rather difficult barriers to be crossed.

We would like it to be the case that, in exploring a bounded area, the robot generates a map whose size is a polynomial function of, say, the number of distinctive places in that bounded area. Most map-learning algorithms can not guarantee this. We can provide such a guarantee for our algorithm, but only by making two rather restrictive assumptions: first, that the robot is able to correctly identify landmarks, and, second, that the robot knows the maximum out degree of any vertex in the induced graph. We believe that, if the robot has information about the spatial distribution of locations satisfying certain perceptible properties and that distribution satisfies certain basic criteria, then it is possible to (probably approximately correctly) learn a map in polynomial time. The idea is that, if a robot knows enough about the spatial distribution of the different types of locations, it should be able to identify and reaquire landmarks with a better than chance probability of success; this degree of competence should be sufficient for efficient map learning. Whether or not this is a reasonable assumption is primarily a function of the environment and the power of the robot's sensors. Our feeling is that there are interesting environments in which currently practical sensors are sufficient for competent map learning.